\documentclass[10pt,twocolumn,letterpaper]{article}

\usepackage{iccv}
\usepackage{times}
\usepackage{epsfig}
\usepackage{graphicx}
\usepackage{amsmath}
\usepackage{amssymb}
\usepackage{booktabs} 
\usepackage{subfigure}
\usepackage{algorithm}
\usepackage{algorithmic}
\usepackage{float}
\usepackage{authblk}

\usepackage[pagebackref=true,breaklinks=true,letterpaper=true,colorlinks,bookmarks=false]{hyperref}

\iccvfinalcopy 

\def\iccvPaperID{2431} 

\begin{document}
	
	\title{Fast Regularity-Constrained Plane Reconstruction}
	
	\author[1]{Yangbin Lin\thanks{This work was supported by the National Science Foundation of China (No. 61701191) and Natural Science Fund of Fujian Province (No. 2018J05108)}}
	\author[1]{Jialian Li}
	\author[2]{Cheng Wang}
	\author[2]{Zhonggui Chen}
	\author[1]{Zongyue Wang}
	\author[2]{Jonathan Li}
	
	\affil[1]{Computer Engineering College, Jimei University}
	\affil[2]{Xiamen University}
	
	
	\maketitle
	
	\begin{abstract}
		Man-made environments typically comprise planar structures that exhibit numerous geometric relationships, such as parallelism, coplanarity, and orthogonality. Making full use of these relationships can considerably improve the robustness of algorithmic plane reconstruction of complex scenes. This research leverages a constraint model requiring minimal prior knowledge to implicitly establish relationships among planes. We introduce a method based on energy minimization to reconstruct the planes consistent with our constraint model. The proposed algorithm is efficient, easily to understand, and simple to implement. The experimental results show that our algorithm successfully reconstructs planes under high percentages of noise and outliers. This is superior to other state-of-the-art regularity-constrained plane reconstruction methods in terms of speed and robustness.
		
	\end{abstract}
	
	\section{Introduction}
	
	With the proliferation of three-dimensional (3D) scanning devices, it is easy to acquire large volumes of 3D point clouds. Point clouds provide a direct and convenient way to describe outdoor and indoor scenes. However, the method is imposing when modeling such massive data. Automatic modeling methods typically produce excessively complex meshes, which are cumbersome for later applications. However, many applications (\eg, simultaneous localization and mapping and level-of-detail generation) do not require overly fine models. Considering that man-made environments contain many planar structures, plane reconstruction remains a suitable choice for 3D-scene description. Furthermore, plane reconstruction or extraction is typically used as a prior step in various tasks, such as architecture modeling~\cite{Arikan2013O}, place recognition~\cite{Fernandez2013Fast}, object recognition~\cite{Oesau2016Object}, and registration~\cite{Zhou2012Robust}.
	
	In man-made environments, planar structures generally conform to one of the following relationships: parallelism, orthogonality, coplanarity, and angular equality. Making full use of these relationships can considerably improve the robustness of the plane reconstruction algorithms for complex scenes~\cite{Monszpart2015RAPter}. However, traditional plane extraction techniques (\eg, region growing, Hough transform~\cite{Xu1990RHT,Borrmann2011Hough}, and random-sample consensus (RANSAC)~\cite{Fischler1981Random, Schnabel2010RANSAC}) do not take advantage of these geometric constraints and suffer from a lack of robustness under high percentages of noise and outliers.
	
	Next, we refer to the task of reconstructing planes from 3D unstructured point clouds while maintaining a geometric relationship between reconstructed planes as regularity-constrained plane reconstruction (RCPR), which is increasingly used in industrial manufacturing and architectural modeling.
	
	A trivial method of RCPR is the detection-then-regularization strategy, which has two steps. In the first step, the method extracts planes without regard to regularity constraints. Regularity is reconstructed in the second step. This method relies heavily on the quality of the extracted planes in the first step and is easily renders local solutions. Sophisticated algorithms, on the other hand, are traditionally time-consuming, hampering practicality.
	
	In this paper, we formulate RCPR as a global $L_0$ gradient minimization problem, proposing a simple but efficient algorithm to solve it. The proposed algorithm solves the problems of simultaneous plane reconstruction and regularization without losing efficiency. In fact, our algorithm is 3-1,000 times faster than extant RCPR algorithms.
	
	\section{Related work}
	
	In this section, we review works immediately related to our problem. We cover the following three main aspects: plane extraction, multimodel fitting, and regularity-constrained fitting.
	
	\paragraph{Plane extraction}
	
	Plane extraction is an old but recurrent problem. It requires many different techniques and has many variants, depending on the tasks and input data. A very common technique is region growing, often employed by more sophisticated algorithms to extract initial planes~\cite{Chauve2010Robust, Monszpart2015RAPter, Oesau2015Planar, Lin2017Facet}. Another popular approach is RANSAC~\cite{Fischler1981Random} and variants (\eg~\cite{Schnabel2010RANSAC, Raguram2013USAC, Korman2018Latent, Zuliani2005MultiRANSAC}). RANSAC is a robust data estimator when facing a high percentage of outliers. However, it was originally designed to cope with a single model and was not built to maintain regularity. The randomized Hough transform technique~\cite{Xu1990RHT} naturally handles multiple structures. However, it cannot cope with a high percentage of outliers~\cite{Magri2014TLinkage}.
	
	Recent planar-structure extraction techniques (\eg, supervoxel segmentation~\cite{Papon2013Voxel, Lin2018Supervoxel} and structural-scale planar-shape detection~\cite{Fang2018Planar}) constrain the scale of planar structures but do not constrain their geometric relationships. 
	
	In summary, traditional plane extraction methods do not aim to maintain regularity. However, they are often the initial step in the RCPR methods.
	
	\paragraph{Multimodel fitting}
	
	The goal of multimodel fitting is to improve the robustness when fitting multiple models to data contaminated by noise and outliers. Multimodel fitting is typically formulated as an optimization problem. For example, Delong \etal~\cite{Delong2012Fast} and Amayo \etal~\cite{Amayo2018Geometric} converted the problem to a multilabeling problem and solved it via $\alpha$-expansion~\cite{Boykov2002Fast} and a primal-dual algorithm \cite{Chambolle2011PrimalDual}, respectively. Magri and Fusiello~\cite{Magri2016Multiple} cast the multimodel fitting problem as a set-coverage problem and solved it with integer linear programming. Moreover, clustering-based methods (\eg J-linkage~\cite{Toldo2008JLinkage} and T-linkage~\cite{Magri2014TLinkage}) and hypergraph-based methods~\cite{Wang2018Searching} have gained the popularity. 
	
	In our application, because of the huge input point cloud, we favor methods having low time complexity. Unfortunately, most multimodel fitting methods do not meet this requirement.
	
	\paragraph{Regularity-constrained fitting}
	
	Regularity-constrained fitting has gained popularity in recent years. Li \etal~\cite{Li2011GlobFit} presented the GlobFit framework, which adopted an iterative detection-then-regularization strategy to fit primitives. It first used RANSAC to locally fit primitives and to detect the geometric relationships among the primitives. Then, it refitted these primitives using the detected constraints. Similarly, Zhou, and Neumann~\cite{Zhou2012Building} adopted the same strategy to model buildings. Departing from GlobFit, Oesau \etal~\cite{Oesau2015Planar} sought to progressively detect regularities rather than iterating between complete detection and regularization. Monszpart \etal~\cite{Monszpart2015RAPter} focused on constraining the angles between pairs of planes. Their method first generated candidate planes with fitting distances. Then, it searched the optimal planes from candidates using mixed-integer linear programming. Although the method achieved good performance with cluttered scenes, it was time-consuming because of the need to solve large-scale mixed-integer programming problems. More recently, Joo \etal~\cite{Joo2018Manhattan} presented a fast branch-and-bound framework to estimate the optimal Manhattan frame.
	
	\section{Motivation and constraint models}
	
	RCPR's regularity depends on the assumptions made of the real world. Here, we refer to this group of assumptions as a constraint model. Moreover, we define the constraint model as a collection of plane sets satisfying a particular constraint. Thus, the task of RCPR can be regarded as finding an element from the constraint model that best fits the given data. A typical constraint model is the Manhattan model~\cite{Coughlan1999Manhattan}.
	
	\paragraph{Manhattan model} A plane set, $\mathcal{P}$, is a Manhattan model if and only if it satisfies
	\begin{equation}
	\angle v_i, v_j \in \{0^\circ, 90^\circ \}, \forall v_i, v_j \in V,
	\end{equation}
	where $V$ is the set of unit normal vectors of $\mathcal{P}$. Because a plane has two orientations, we specify that the angle between the normal vector in $V$ and positive z-axis is no more than $90^\circ$.
	
	In practice, the Manhattan model is a very strong constraint. It can simplify a few tough problems, such as line clustering and vanishing-point estimation in two dimensional (2D) images~\cite{Bazin2012Globally}. However, the real world usually does not conform to the Manhattan model. Therefore, Straub \etal~\cite{Straub2014MMF} proposed a mixture of Manhattan frames. The idea was to utilize multiple Manhattan frames (MMF) to represent complex scenes.
	
	\paragraph{Multiple Manhattan frames model} 
	A plane set $\mathcal{P}$, is an $n$-MMF model if it can be divided into $n$ non-overlapping subsets, with each subset being a Manhattan model. 
	
	The limitation of Manhattan and MMF models is obvious. For example, they only consider parallelism and orthogonality. RAPTER~\cite{Monszpart2015RAPter} adopted a constraint model that considered more angular constraints. Because RAPTER's constraint model is difficult to formulate, we only consider its variant, which we call the generalized Manhattan model.
	
	\paragraph{Generalized Manhattan model}
	A plane set, $\mathcal{P}$, is a generalized Manhattan model, if and only if it satisfies
	\begin{equation}
	\angle v_i, v_j \in \mathcal{S}, \forall v_i, v_j \in V,
	\end{equation}
	where $S$ is a user-defined set of constraint angles. It is easy to note that, if $S = \{0^\circ, 90^\circ \}$, the generalized Manhattan model is equivalent to the Manhattan model.
	
\begin{figure}[ht]
		\centering
		\subfigure[] {	
			\includegraphics[width=0.51in]{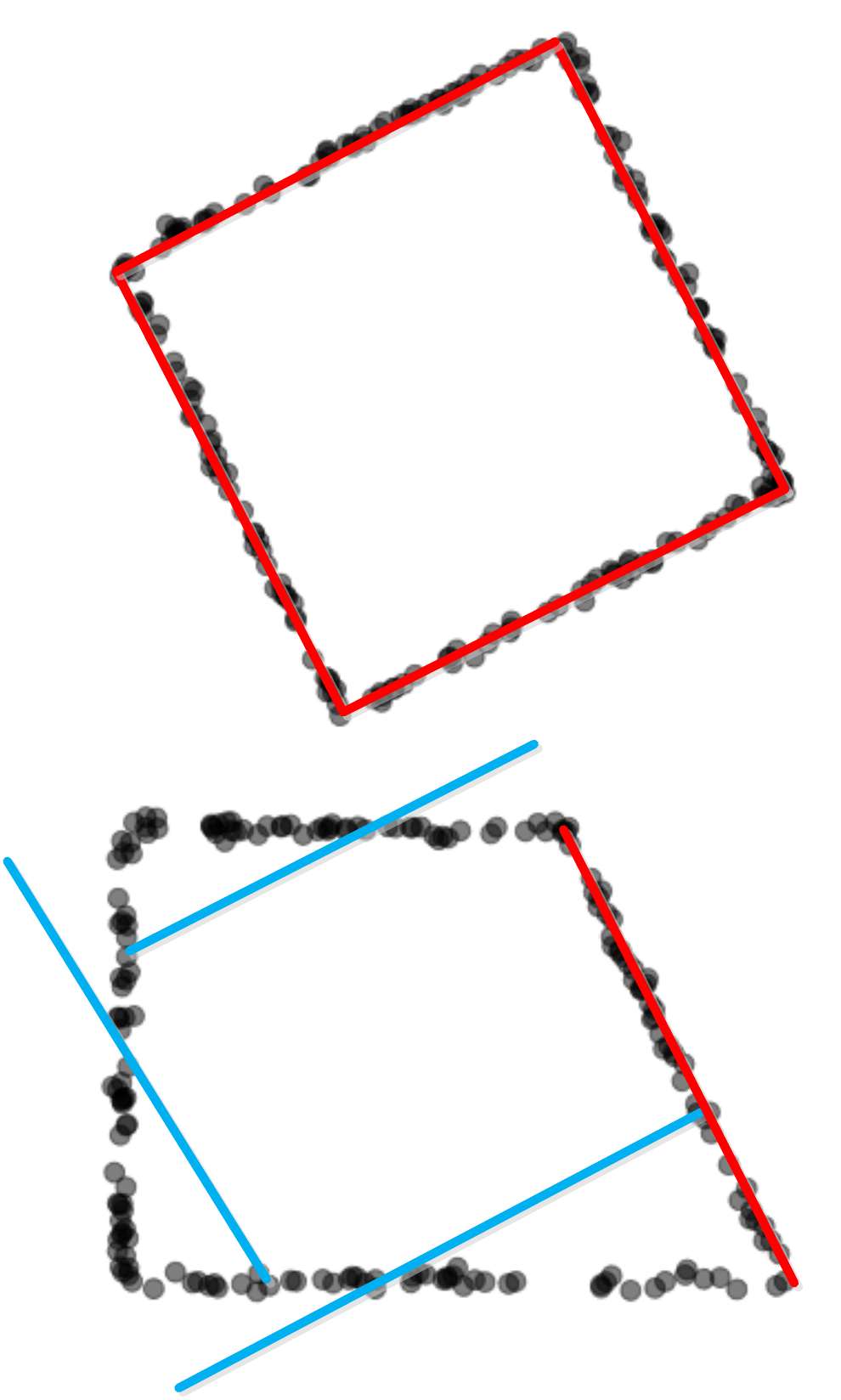} 
		}
		\hspace{0.1in}
		\subfigure[] {	
			\includegraphics[width=0.5in]{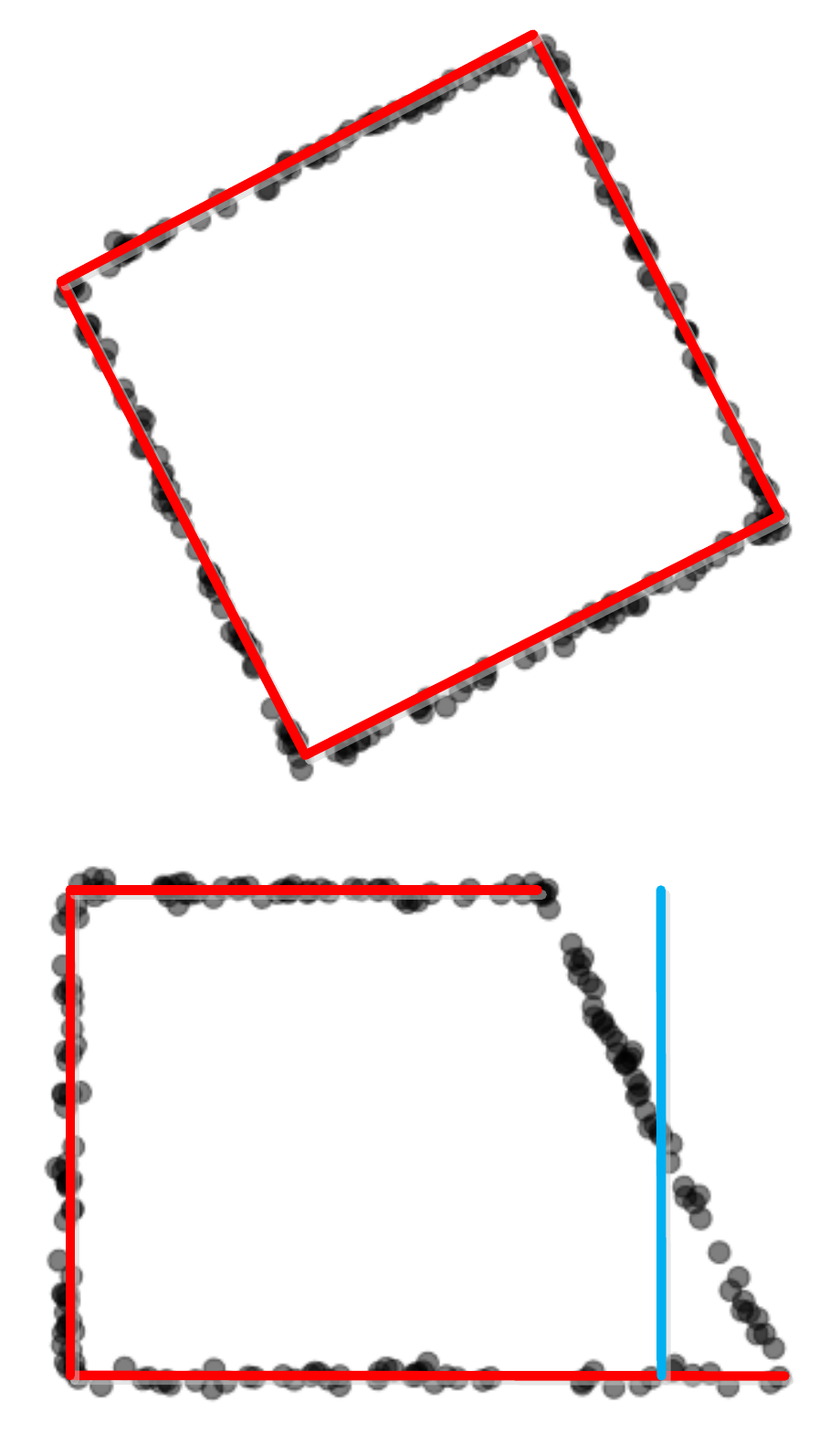} 
		}
		\hspace{0.1in}
		\subfigure[] {	
			\includegraphics[width=0.5in]{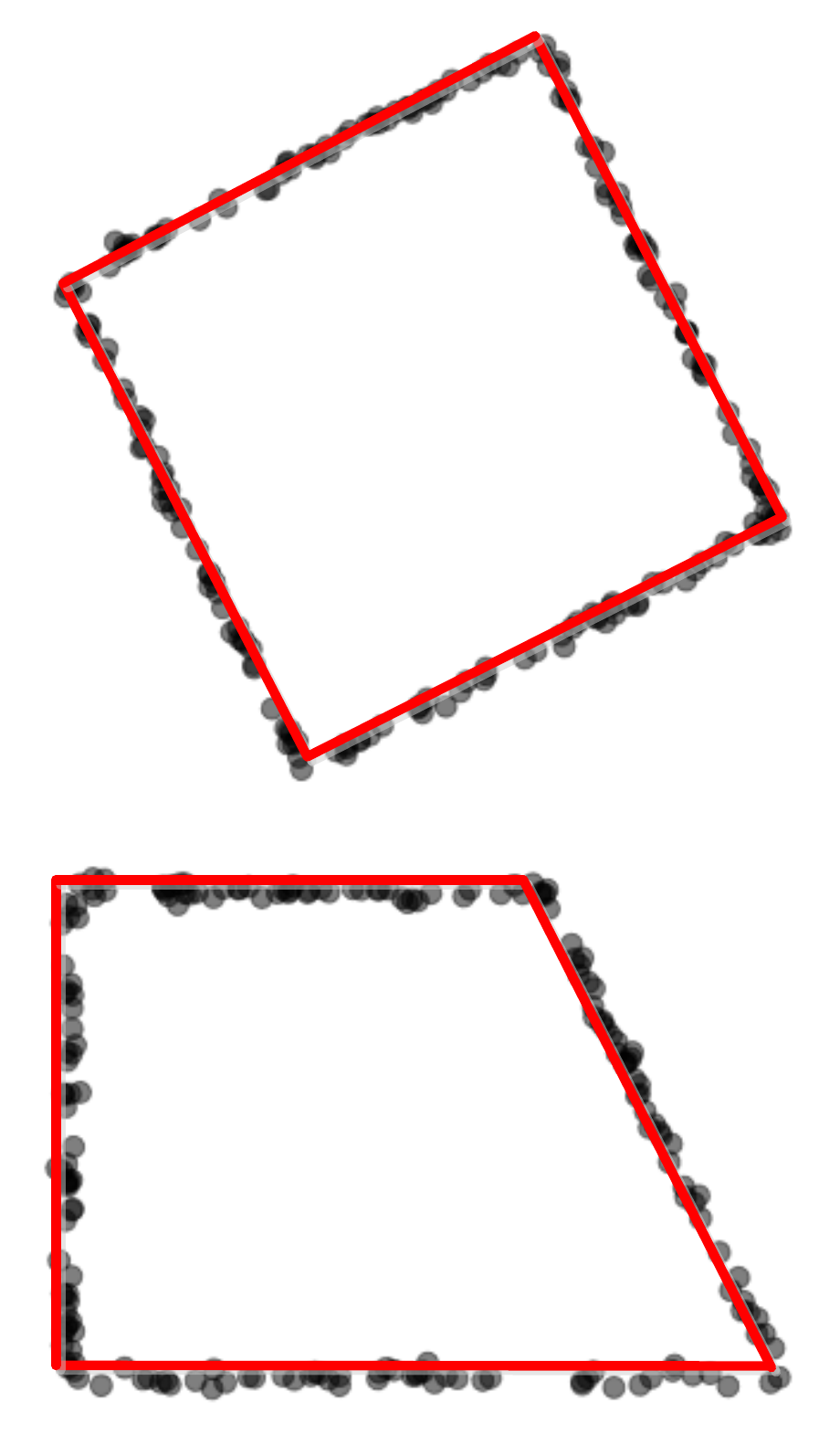} 
		}
		\hspace{0.1in}
		\subfigure[] {	
			\includegraphics[width=0.5in]{constraint_models_3} 
		}
		
		\caption{Line-fitting results using four constraint models: (a) Manhattan model; (b) MMF model, where $n$, the parameter that controls the number of Manhattan frames, is set to 2; (c) generalized Manhattan model, where the set of constraint angles, $S$, is set to $\{0^\circ, 25.565^\circ, 63.435^\circ, 90^\circ, 116.565^\circ, 153.435^\circ \}$; (d) our directional constraint model, where the number of different directions, $m$, is set to 4.}
		\label{fig:constraint_models}
	\end{figure}

	A comparison of these constraint models on a line-fitting task is illustrated in Fig.~\ref{fig:constraint_models}. We notice that the Manhattan model cannot fit this simple data well. The MMF obtains a better result but still misses one line. The generalized Manhattan model can fit the data well. However, it should introduce more angle constraints. For example, in Fig.~\ref{fig:constraint_models}, the set of constraint angles should be set to $\{0^\circ, 25.565^\circ, 63.435^\circ, 90^\circ, 116.565^\circ, 153.435^\circ \}$. Usually, it is difficult to know the exact constraint angles. Furthermore, more constraint angles considerably increase the complexity of regularization.
	
	We consider, therefore, using implicit angle constraints instead of explicit angle constraints. Specifically, we introduce a parameter, $m$, to constrain the number of different normal vectors of the reconstruction planes. Our motivation is based on a simple observation: In man-made scenes, the number of planes, $|\mathcal{P}|$, is always much more than the number of different normal vectors, $|V|$. For example, in Fig.~\ref{fig:constraint_models}, we have eight lines, but only four different directions (i.e., normal vectors). We refer to our constraint model as a directional constraint (DC) model, defined as follows.
	
	\paragraph{Directional constraint model}
	Given a plane set $\mathcal{P}$, we denote $V$ as the set of different normal vectors of $\mathcal{P}$. $\mathcal{P}$ is a $m$-DC model, if and only if it satisfies
	\begin{equation}
	|V| <= m.
	\end{equation}
	
	Apart from the other three models, the DC model cannot maintain orthogonality. However, it gains the advantage of fewer priori parameters. In fact, the DC model only needs one priori parameter, $m$, (i.e., the number of different normals). An example is given in Fig.~\ref{fig:constraint_models} (d). There are eight lines but only four different directions. Therefore, we can set $m=4$ to obtain the correct fitting result. Moreover, by constraining the number of normal vectors, we implicitly establish the connection between planes, which helps obtain better fitting results under noise constraints and outliers (See Fig.~\ref{fig:comparison_2d}). 
	
	In the next section, we discuss how to reconstruct a set of planes satisfying the DC model.
	
	\section {Methodology}
	
	\subsection{Problem formulation}
	Given a point cloud, $D$, equipped with normal vectors, $I$, our task is to find a plane set, $\mathcal{P}$, that best fits the input points while satisfying the DC model. If the input point cloud has no normal vector information, we use the principal component analysis algorithm to estimate their normal vectors.
	
	Our method comprises two steps. First, we reconstruct the normal vector for each point while maintaining the constraints of DC model. Second, the desired planes can be easily obtained by grouping nearby points having the same reconstructed normal. Next, we focus on the first step, because the second step is easily implemented. More specifically, we consider the following problem:
	\begin{equation}
	\label{equ:problem1}
	\min_V E(I, V) \quad s.t. \ |V| <= m,
	\end{equation}
	where, $V$ is the set of reconstructed normal vectors; $E$ is the energy that measures how well $V$ fits $I$. Here, we adopt $L_0$ energy for $E$, owing to its performance under high noise. Moreover, the constraint term, $|V|<=m$, can be relaxed by introducing a regularization parameter $\lambda_g$. Therefore, Eq.~\ref{equ:problem1} is rewritten as
	\begin{equation}
	\label{equ:problem2}
	\min_V \sum_i^M\left(||V_{z_i} - I_i||^2 + \lambda_l \sum_{j\in \mathcal{N}_i}||V_{z_i} - V_{z_j}||_0\right) + \lambda_g |V|,
	\end{equation}
	where $M$ is the number of points; $\mathcal{N}_i$ is the neighboring set of the $i$-th point; and $z_i$ is the index of $V$ for the $i$-th point. Note that Eq.~\ref{equ:problem2} can also satisfy the DC model if we choose the appropriate $\lambda_g$ value.
	
	In Eq.~\ref{equ:problem2}, the first term, $||V_{z_i}-I_i||^2$, is used to guide the output normal vectors as close as possible to the input. The second term, $\sum_{j\in \mathcal{N}_i}||V_{z_i} - V_{z_j}||_0$, is used to constrain the sparsity in the local sense. It guides the normal vector of each point close to its neighboring normals. The third term, $|V|$, is used to constrain the global sparsity. Two parameters, $\lambda_l$ and $\lambda_g$, are used to balance the influence of two constraint terms. Eq.~\ref{equ:problem2} can be further divided into the following two subproblems: 
	
	\paragraph{Subproblem 1}
	\begin{equation}
	\label{equ:subproblem1}
	E_1(I, \lambda) = \min_V \sum_i^M\left(||V_{z_i}-I_i||^2 + \lambda \sum_{j\in \mathcal{N}_i}||V_{z_i} - V_{z_j}||_0\right).
	\end{equation}
	
	\paragraph{Subproblem 2}
	\begin{equation}
	\label{equ:subproblem2_0}
	E_2(I, \lambda) = \min_V \sum_i^M||V_{z_i}-I_i||^2 + \lambda |V|.
	\end{equation}
	If we limit $V \subseteq I$, then Eq.~\ref{equ:subproblem2_0} can be rewritten as:
	\begin{equation}
	\label{equ:subproblem2}
	E_2(I, \lambda) = \min_{V\subseteq I} \sum_i^M \min_{v \in V}||v-I_i||^2 + \lambda |V|.
	\end{equation}
	
	We notice that Eq.~\ref{equ:subproblem1} is an $L_0$ gradient minimization problem and that Eq.~\ref{equ:subproblem2} is a subset selection problem. This provides a simple notion: we can optimize Eq.~\ref{equ:problem2} by alternately solving two subproblems. By reviewing the state-of-the-art $L_0$ minimization~\cite{Nguyen2015Fast} and subset selection techniques~\cite{Elhamifar2017Online}, we can provide a simple but efficient algorithm called, "Global-$L_0$", to solve the problem.
	
	\subsection{Global-$L_0$ algorithm}
	The framework of the Global-$L_0$ algorithm is given in Algorithm~\ref{algorithm1}, comprising the following two steps.
	\begin{algorithm}
		\caption{Global-$L_0$ algorithm}
		\textbf{Input:} The input point cloud, $D$; the normal vectors, $I$; the neighboring set, $\mathcal{N}$; the local and global regularization parameter, $\lambda_l$ and $\lambda_g$. \\
		\textbf{Output:} the constrained reconstruction normal vectors $V$.
		
		\begin{algorithmic}[1]
			\label{algorithm1}
			\STATE $\lambda = \mathrm{median}\{\min_{p_i\in D}||I_i - I_{\mathcal{N}_i}||\}$
			\STATE $V = I$
			\REPEAT 
			\STATE $L_0$ minimization: 
			$V^o \gets \underset{V^o}{\mathrm{argmin}} \ E_1(V, \lambda)$
			
			\STATE Subset selection: 
			$V^n \gets \underset{V^n}{\mathrm{argmin}} \ E_2(V^o, \lambda\cdot \lambda_g)$
			
			\STATE $V = V^n$
			\STATE $\lambda \gets 2\lambda$
			\UNTIL {$\lambda > \lambda_l$}
			\RETURN $V$
		\end{algorithmic}
	\end{algorithm}
	
	\paragraph{$L_0$ minimization}
	To solve subproblem~\ref{equ:subproblem1}, we adopt the algorithm based on region fusion~\cite{Nguyen2015Fast}. Here, we only need to perform one iteration with the fixed $\lambda$ value. The time complexity is $O(M)$.
	
	\paragraph{Subset selection}
	To solve subproblem~\ref{equ:subproblem2}, we adopt greedy unconstrained submodular optimization. Let, $F(V) = -E_2(V^o, \lambda)$. $F(V)$ is a submodular function satisfying $F(X \cup \{x\}) - F(X) \geq F(Y \cup \{x\}) - F(Y)$ for any $X \subseteq Y \subseteq V \setminus \{x\}$. Therefore, $F(V)$ is an unconstrained submodular maximization (USM) problem. We can use the linear-time algorithm proposed in \cite{Buchbinder2012A} to solve it. 
	
	However, the complexity of using Algorithm~\cite{Buchbinder2012A} directly is $O(M^2)$, which is unacceptable. We notice that, after performing region-fusion-based $L_0$ minimization, the normal vectors, $V^o$, are divided into several disjoint regions. Each region shares the same normal vector. Therefore, $F(V)$ can be simplified to
	\begin{equation}
	F(V) = -\min_{V\subseteq V^\mathcal{S}}\sum_{v_j\in V} \min_{s_i\in \mathcal{S}} ||\hat{s_i} - v_j||^2 |s_i| + \lambda |V|,
	\end{equation}
	where $\mathcal{S}$ is the set of regions; $\hat{s_i}$ is the normal vector of the $i$-th region; and $V^\mathcal{S}$ is the set of normal vectors consisting of $\hat{s_i}$, $i \in [1, |\mathcal{S}|]$.
	
	Additionally, we can ignore the regions whose number of points is less than a threshold, $\tau$. $\tau$ can be explained as the minimal support points for a plane (i.e., the filtered regions are treated as outliers). Thus, the time complexity of our subset selection step is reduced to $O(\min(M / \tau, |\mathcal{S}|)^2)$. In practice, it can assume $O(\min(M / \tau, |\mathcal{S}|)^2) < O(M)$.
	
	More details are described in Algorithm~\ref{algorithm3}.
	
	\begin{algorithm}
		\caption{Subset selection}
		\textbf{Input:} The disjoint region set $\mathcal{S}$; each region shares the same normal vector $\hat{s_i}$. \\
		\textbf{Output:} the selected normal vectors, $V$.
		
		\begin{algorithmic}[1]
			\label{algorithm3}
			\STATE Sort $\mathcal{S}$ in descending order according to their number of points.
			
			\STATE Initialize: $V \gets \emptyset$; $X\gets\emptyset$; $Y\gets \mathcal{S}$
			\FOR {$i = 1,...,|\mathcal{S}|$ and $|\mathcal{S}_i| >= \tau$}
			\STATE $a \gets F(X \cup \{\mathcal{S}_i\}) - F(X)$
			\STATE $b \gets F(Y \setminus \{\mathcal{S}_i\}) - F(Y)$
			\IF {$a \geq b$}
			\STATE $X\gets X\cup \{S_i\}$
			\STATE $V\gets V\cup \{\hat{s_i}\}$
			\ELSE
			\STATE $Y\gets Y \setminus \{\mathcal{S}_i\}$
			\ENDIF
			\ENDFOR
			\RETURN $V$
		\end{algorithmic}
	\end{algorithm}
	
	Algorithm~\ref{algorithm3} has a 1/3 approximation guarantee (i.e., the energy of its solution is always more than 1/3 times the optimal energy). The randomized version~\cite{Buchbinder2012A} has a better approximate rate of 1/2. However, we choose the deterministic algorithm, because it is more stable. Furthermore, with a heuristic sequence (i.e., descending order by cardinality), our solution is usually better than the theoretical upper bound.
	
	The Global-$L_0$ algorithm can be seen as an extension of $L_0$ gradient minimization algorithms. The traditional $L_0$ gradient minimization methods only consider $L_0$ regularity between local neighbors. Our algorithm introduces global regularity by limiting the number of different normal vectors. It establishes implicit relationships between the disjoint regions. In Section~\ref{sec:Experiments}, we demonstrate that these global relationships are very useful, especially in man-made environments.
	
	\paragraph{Plane reconstruction}
	For a post-processing, the constrained planes can be constructed from disjoint regions. Additionally, the normal vector of each plane is set to the reconstruction normal vector of the corresponding region.
	
	\subsection{Difference to the multilabeling problem}
	Eq.~\ref{equ:problem2} looks like the multilabeling problem of~\cite{Delong2012Fast,Amayo2018Geometric}. However, they are quite different. First, the energy formulation in~\cite{Delong2012Fast,Amayo2018Geometric} only constrained the number of models (planes in our context). In contrast, Eq.~\ref{equ:problem1} constrains the number of plane normals, because, it constrains the size of the label set. Second, in the multi-labeling problems, the label set, $L$ (or $V$ in our formulation), is usually known. For example, Pearl~\cite{Delong2012Fast} must first compute an initial set of models or labels via sampling. Our algorithm is designed to solve the problem without an initial label set. In other words, the size of the initial label set in our problem is much larger than the label size in multilabeling problems. Therefore, an efficient algorithm is desired.
	
	\section{Experiments}
	\label{sec:Experiments}
	
	Our method was implemented in C++ and run on Linux Ubuntu 16.04 with one core of Intel i7 CPU (2.5 GHz) and 8-GB memory.
	
	\subsection {Parameter setting}
	There are four parameters adopted in our method. They are $\tau$, the minimal number of points that support a plane; $K$, the number of neighbors for each point; $\lambda_l$, the local regularization for normal vectors in $L_0$ minimization; and $\lambda_g$, the global regularization for normal vectors. $\tau$ and $K$ are common parameters, and their settings have been fully studied. Here, we set $\tau = 100$ and $K = 10$ for the following experiments. 
	
	\begin{figure}[ht]
		\centering
		\includegraphics[width=\linewidth]{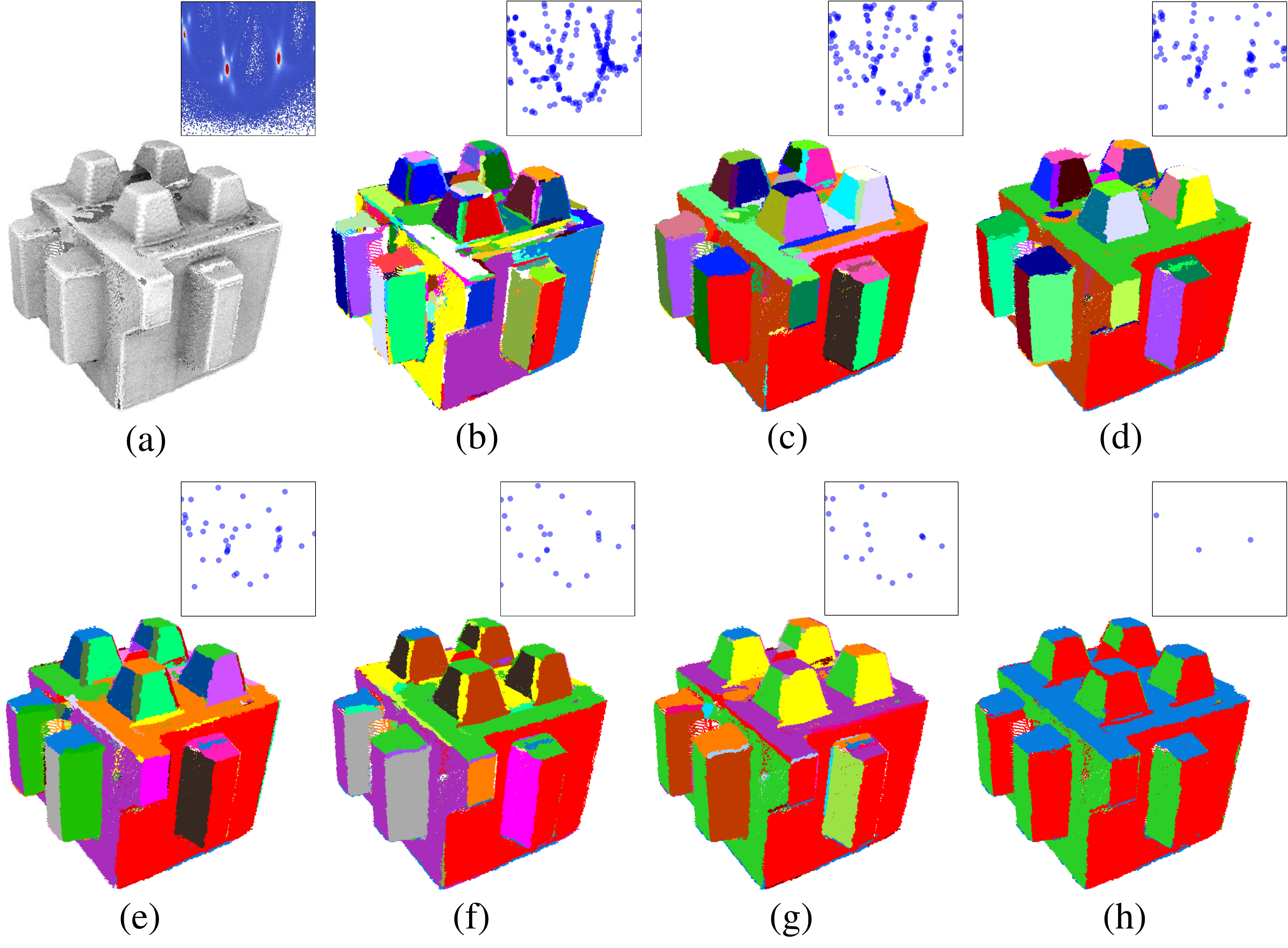}
		
		\caption{Influence of different values of $\lambda_l$ and $\lambda_g$ on the results: (a) original point cloud; (b--d) $\lambda_g = 0$, $\lambda_l = 0.1, 0.5, 1$, respectively; (e--h) $\lambda_l = 0.5$, $\lambda_g = 100, 500, 1,000, 10,000$, respectively. The black boxes indicate a normal vector distribution. }
		\label{fig:foam_box}
	\end{figure}
	
	Fig.~\ref{fig:foam_box} demonstrates the effects of $\lambda_l$ and $\lambda_g$. We use the same color to draw the points with the same normal vector. In Fig.~\ref{fig:foam_box}(b--d), we first set $\lambda_g = 0$ and increase the value of $\lambda_l$. One can observe that, as the $\lambda_l$ increases, the number of regions decreases. However, the normal vectors for each region are different (i.e., $|V|$ equals the number of regions). In Fig.~\ref{fig:foam_box}(e-h), we set $\lambda_l=0.5$ and increase the value of $\lambda_g$. The number of regions and the different normal vectors decrease simultaneously. When $\lambda_g$ reaches 10,000, only the three most frequently normal are preserved, satisfying the Manhattan model.
	
	In practice, for data having high noise, we choose higher values for $\lambda_l$ and $\lambda_g$. One can also simply set $\lambda_l$ equal to the one in \cite{Nguyen2015Fast}. Additionally, if the user has a prior knowledge of the scene (the value of $m$), then one can choose an appropriate $\lambda_g$ value.
	
	\subsection{2D line fitting}
	We first study the robustness of the proposed method on the 2D line-fitting problem with respect to seq-RANSAC, J-linkage~\cite{Toldo2008JLinkage}, T-linkage~\cite{Magri2014TLinkage}, and Pearl~\cite{Delong2012Fast}. We implement seq-RANSAC, J-linkage, and T-linkage ourselves, whereas the implementation of Pearl is obtained from \cite{PearlCode}.
	
\begin{figure*}[ht]
		\centering
		\includegraphics[width=\linewidth]{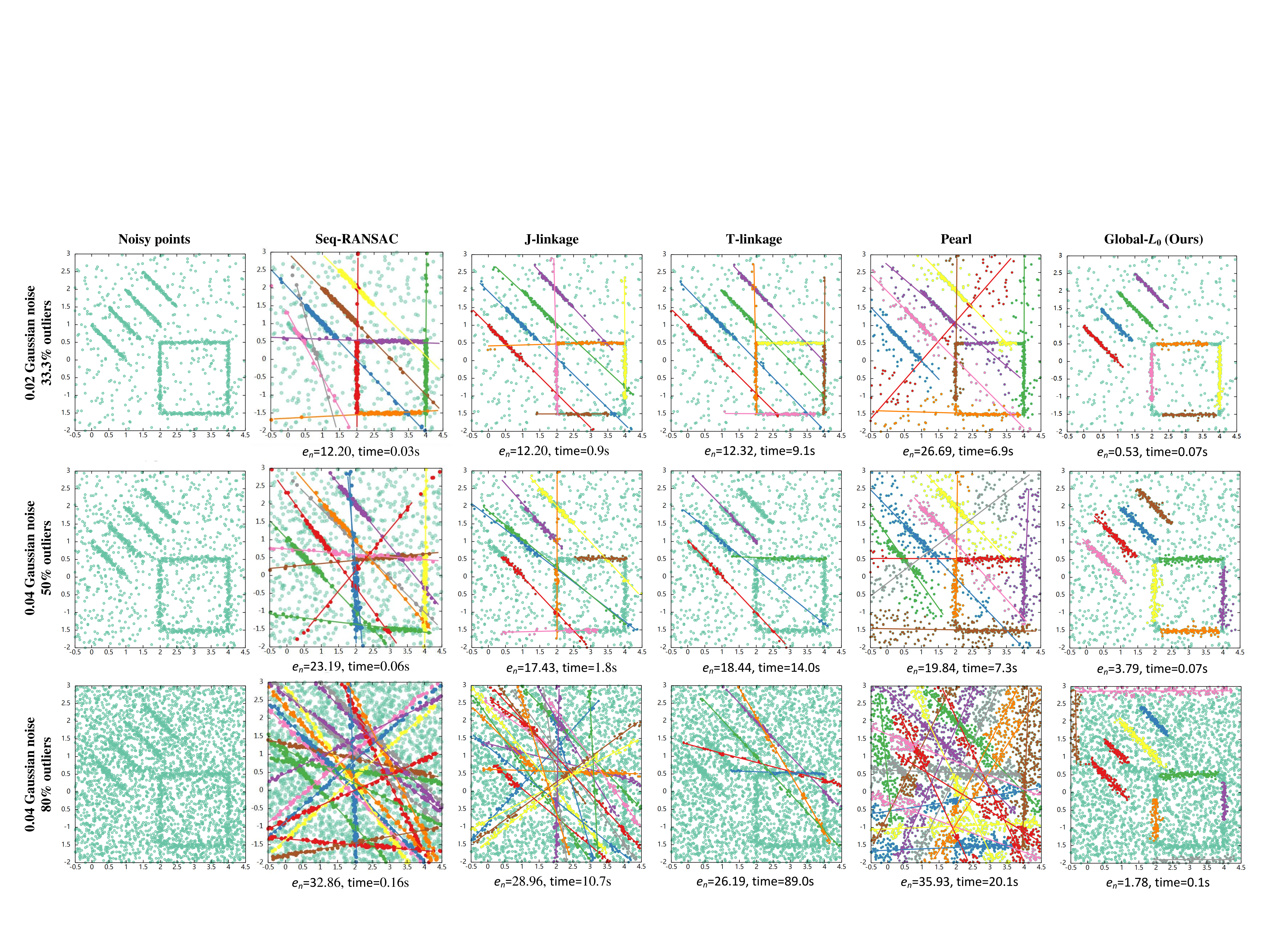}
		\caption{Comparison of line-fitting results on 2D synthesized data with different levels of Gaussian noise and outliers.}
		\label{fig:comparison_2d}
	\end{figure*}

	As shown in Fig.~\ref{fig:comparison_2d}, we generate eight lines, four of which are parallel, and four of which form a square. Each line comprises of 100 inliers, contaminated by Gaussian noise and outliers of different percentages. When we say 50\% outliers, it means that the number of outliers is half of the entire set. Note that the ground truth lines only have three different directions. Our method can benefit from this constraint. 
	
	The results are collected in Fig.~\ref{fig:comparison_2d}, where $e_n$ is the root-mean-square (RMS) error of the angles between the reconstruction of normal vectors and ground truth normals. It is defined as
	\begin{equation}
	e_n = \sqrt{\frac{\sum_{p \in \mathcal{G}}(\angle(\hat{n}_p, n_p))^2}{|\mathcal{G}|}},
	\end{equation}
	where $\mathcal{G}$ represents the ground truth inliers.
	
	As noise and outliers increase, Seq-RANSAC, J-linkage, T-linkage, and Pearl miss the ground-truth line segments, and generate more false positives. J-linkage and T-linkage aggregate the preference set without exploiting the regularity of the data. Moreover, J-linkage and T-linkage rely on the selection of the initial set. The tendency of decreasing the percentage of inliers affects the performance of J-linkage and T-linkage. Pearl solves the line-fitting problem by presenting an optimization framework. However, it only constrains the number of fitted lines and does not constrain the number of directions. It also relies on the selection of the initial set.
	
	Our method achieves the best $e_n$ in all situations. The reason can be ascribed to the constraint on the number of directions and the independence of the initial set. Note that, the number of directions is a non-local constraint, which similar to non-local denoising methods~\cite{Buades2005NonLocalDenoising}. When outliers reach 80\% (20\% inliers), our method generates two false positives and misses one line segment. However, the directions of the fitted lines are correct, and the number of different directions (including false positives) remains three. Moreover, our method has the best time performance, significantly faster than the other three methods.
	
	\subsection{Plane reconstruction on 3D manual point sets}
	We further study the robustness of the proposed method on 3D data with respect to RAPTER~\cite{Monszpart2015RAPter}, piecewise planar surface 
	reconstruction (PPSR)~\cite{Chauve2010Robust}, and Pearl~\cite{Delong2012Fast}. We obtain the implementation of RAPTER from~\cite{RAPteCode}. Here, we use the region-growing technique of~\cite{Chauve2010Robust} to extract the initial planes for Pearl, leading to better reconstruction results.
	
\begin{figure}[ht]
		\centering
		\includegraphics[width=3in]{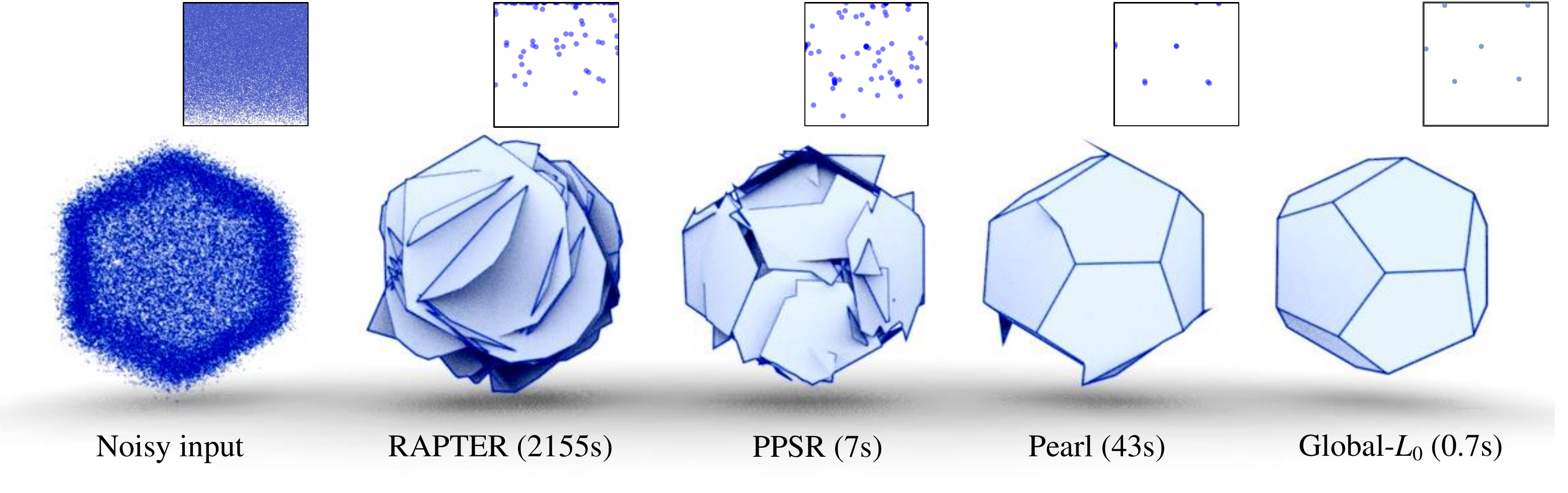}
		\caption{Comparison results between our method and various plane reconstruction methods on a noisy dodecahedron. The mesh models are constructed by point-set structuring~\cite{Lafarge2013Surface} based on the reconstructed planes. The normal vectors of the obtained planes are shown in black boxes. In parentheses, we provide the running time (s) of each method.}
		\label{fig:dodecahedron_noise}
	\end{figure}

	\paragraph{Noisy dodecahedron}
	We first demonstrate the performance of these methods on a noisy 3D point set. As shown in Fig.~\ref{fig:dodecahedron_noise}, we sampled 100,000 points from a dodecahedron and perturbed each point by adding $0.02l$ Gaussian noise, where $l$ is the diagonal length of the axis-aligned bounding box of the original point cloud.
	
	RAPTER is susceptible to Gaussian noise, and, because it relies on the initial planes obtained from region growing, its results will be less than satisfactory when region growing fails. PPSR improves the performance of traditional region growing by carefully selecting seed points. However, it still generates too many redundant planes. Pearl~\cite{Delong2012Fast} obtains a much better result but loses accuracy at the boundaries. Benefiting from the robustness of $L_0$ optimization and parallel faces constraints, our method estimates the correct model. Furthermore, our method is much faster than the others.
	
	\paragraph{Empire State Building with outliers}
	We next test our method with RAPTER on the \emph{Empire State Building} with different level of outliers. The ground truth model is downloaded from~\cite{SurfaceReconstructionData}. We sample 1,000,000 points from ground-truth model and add 250,000 and 1,000,000 outliers, respectively.
	
	In addition to $e_n$, we evaluate the accuracy of the reconstruction planes by computing the RMS of the distance from the ground-truth points, $\mathcal{G}$, to the reconstruction planes, $\mathcal{P}$, i.e.,
	\begin{equation}
	\label{equ:ep}
	e_p = \sqrt{\frac{\sum_{p \in \mathcal{G}}(dist(p, \mathcal{P}_{L(p)})^2}{|\mathcal{G}|}},
	\end{equation}
	where, $L(p)$ denotes the index of the plane closest to $p$.
	
	$e_p$ in Eq.~\ref{equ:ep} is used to penalize the insufficient reconstruction of planes. However, it does not penalize redundant planes, which is critical for point clouds with outliers. Therefore, for each inlier point, $p$, we calculate its projection point, $p'$, on the reconstruction planes. Then, we evaluate the distance from $p'$ to the nearest point using ground truth points, i.e.,
	\begin{equation}
	\label{equ:eq}
	e_q = \sqrt{\frac{\sum_{p \in P \land L(p) \neq Outlier}(\min_{q\in \mathcal{G}} dist(p', q))^2}{|P|}}.
	\end{equation}
	
	Note that $e_q$ can penalize the inaccurate and redundant reconstruction of planes simultaneously.
	
	\begin{figure}[ht]
		\centering
		\includegraphics[width=3in]{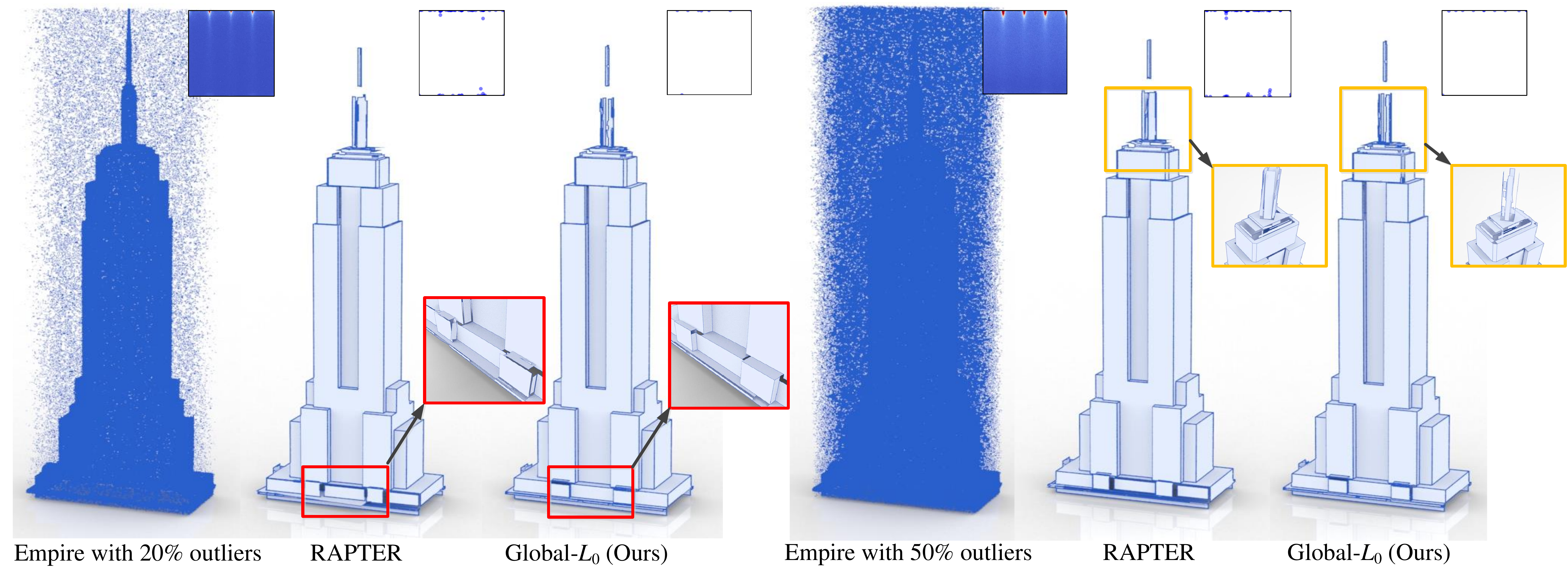}
		\caption{Comparison of our method and RAPTER on "Empire state building" with 20\% and 50\% outliers. The black boxes show the distribution of normal vectors.}
		\label{fig:empire}
	\end{figure}
	
	As shown in the black boxes in Fig.~\ref{fig:empire}, the normal vector distribution of the original point clouds is very noisy. Nevertheless, both methods achieved visually good plane reconstruction results. As a defect, neither method can produce a complete plane reconstruction at the top part of the building (see close-up views in the yellow boxes). The reason is that the small plane is more susceptible to outliers, resulting in incorrect reconstruction planes. Despite that, we can observe that RAPTER produces several planes with inaccurate directions; whereas our method produces relatively few but more accurate planes. In the close-up views in the red boxes, we also observe that our method outperforms RAPTER in details. 
	
	\begin{table}
		\caption{Comparisons on the Empire State Building in terms of $e_n$, $e_p$, $e_q$, and running time. The first two rows are the comparison results for the data with 20\% outliers and the next two rows are the results for the data with 50\% outliers.}
		\label{tab:empire}
		\begin{center}
			\begin{tabular}{|c|c|c|c|c|c|}
				\hline
				Method & \#Planes & $e_n$ & $e_p$ & $e_q$ & Time \\
				\hline
				
				RAPTER & 195 & 3.5 & \textbf{1.1e-4} & 6.5e-5 & 1.3e5 \\ 
				
				Ours & 171 & \textbf{1.8} & 1.5e-4 & \textbf{3.8e-5} & \textbf{55} \\
				\hline
				
				RAPTER & 211 & 3.5 & \textbf{1.1e-4} & 7.8e-5 & 1.5e5 \\ 
				Ours & 178 & \textbf{1.8} & 1.5e-4 & \textbf{3.8e-5} & \textbf{38} \\
				\hline
			\end{tabular}
		\end{center}
	\end{table}
	
	Quantitative comparison results are summarized in Table~\ref{tab:empire}. Compared to RAPTER, our method achieves better $e_n$ and $e_q$ but worse $e_p$. The normal vectors reconstructed by our method are significantly more accurate than RAPTER. This is also indicated in the black boxes of Fig.~\ref{fig:empire}. Furthermore, RAPTER is very time consuming, because it needs to solve a large-scale mixed-integer programming system. In contrast, our method is approximately 2,000 times faster.
	
	\subsection{Plane reconstruction on real-world point clouds}
	To quantitatively evaluate the performance of the proposed method in practice, two real-world dataset are considered. In the following experiments, we set $\lambda_l = 10$ and $\lambda_g = 1,000$. The parameters for sequential RANSAC and Pearl are fine-tuned.
	
	\begin{figure*}[htb]
		\centering
		\includegraphics[width=6in]{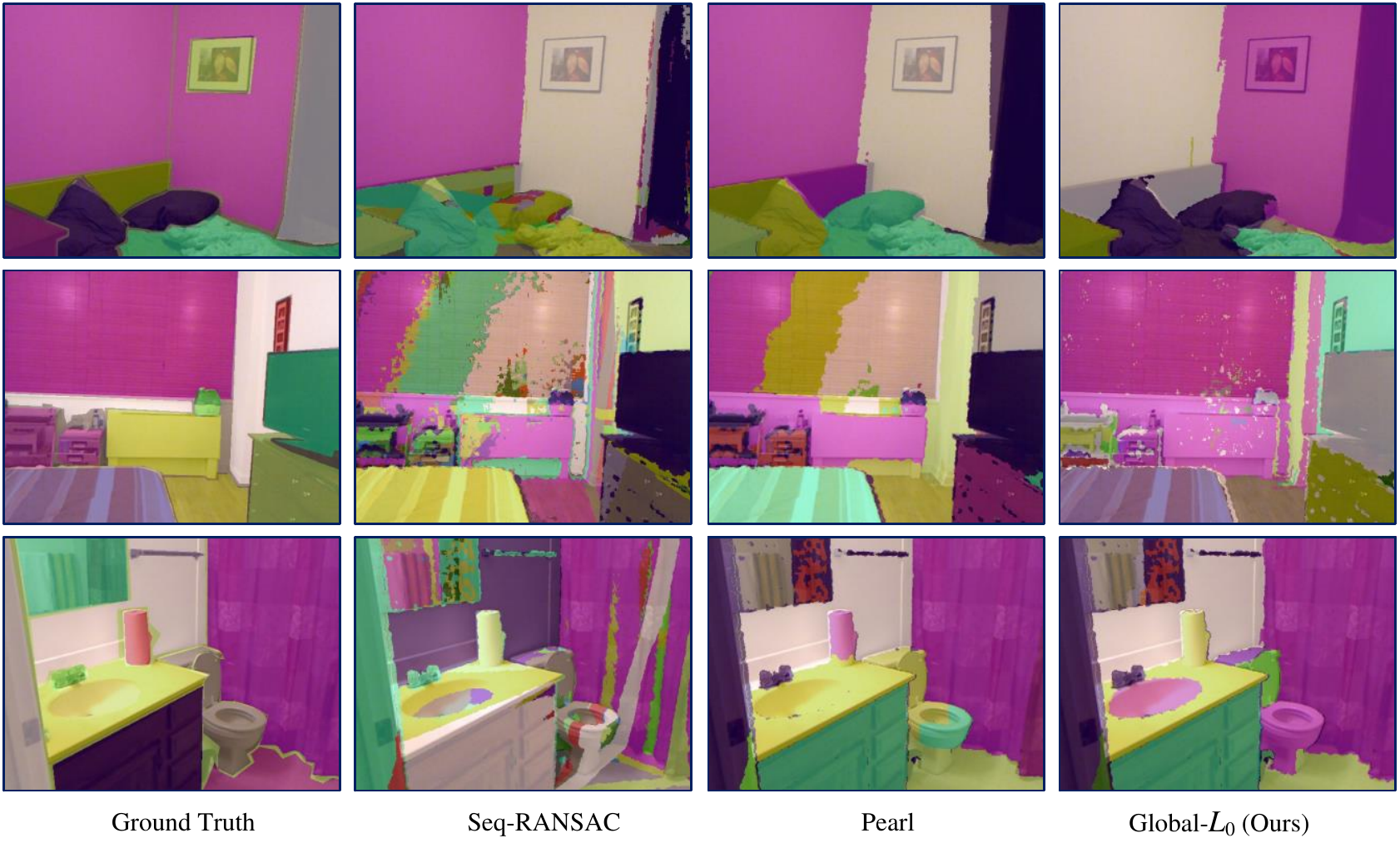}
		\caption{Plane segmentation results using RGBD data. Points on the same plane are drawn in the same color.}
		\label{fig:RGBD}
	\end{figure*}
	
	We first test our method on the NYU-depth dataset \cite{Silberman2012Indoor}. This dataset was originally designed for object detection. We use it to quantitatively evaluate the performance of plane segmentation. A subset of 218 images, mainly containing planar objects, is selected from the dataset. The subset includes kitchen, bedroom, and bathroom scenes. Scenes containing too many non-planar objects, such as offices and stores, are excluded, because our method is not designed for non-planar scenes. Two standard segmentation metrics, global consistency error (GCE), and local consistency error (LCE) from~\cite{Mark2009An}, are introduced here for evaluation.
	
	\begin{table}
		\caption{Plane segmentation comparison results on RGBD dataset. The time is in seconds and averaged over the images.}
		\label{tab:RGBD}
		\begin{center}
			\begin{tabular}{|c|c|c|c|}
				\hline
				Method     		& GCE  & LCE  & Time \\
				\hline
				seq-RANSAC 		& 0.44 & 0.36 & \textbf{7.98} \\
				Pearl      		& 0.36 & 0.28 & 60.11 \\
				Gobal-$L_0$ (Ours) & \textbf{0.33} & \textbf{0.24} & 21.63 \\   
				\hline
			\end{tabular}
		\end{center}
	\end{table}
	
	The quantitative results are summarized in Table~\ref{tab:RGBD}, where two typical scenes are shown in Fig.~\ref{fig:RGBD}. It can be observed that our method is significantly superior to sequential RANSAC and Pearl in terms of visual and quantitative results.
	
	We further test our method on indoor laser-scanning point clouds \footnote{\url{https://www.ifi.uzh.ch/en/vmml/research/datasets.html}}, including five datasets: UZH IfI (12 data), UZH Irchel (10 data), ETH (18 data), Rooms detection datasets (9 data), and Full 3D (8 data). Owing to the memory limitations of PEARL, the input data was down-sampled by performing Poisson-disk sampling \cite{Bridson2007PoissonDisk} using a resolution $r=0.02m$.
	
	\begin{figure}[htb]
		\centering
		\subfigure[$e_p$] {\includegraphics[width=1in]{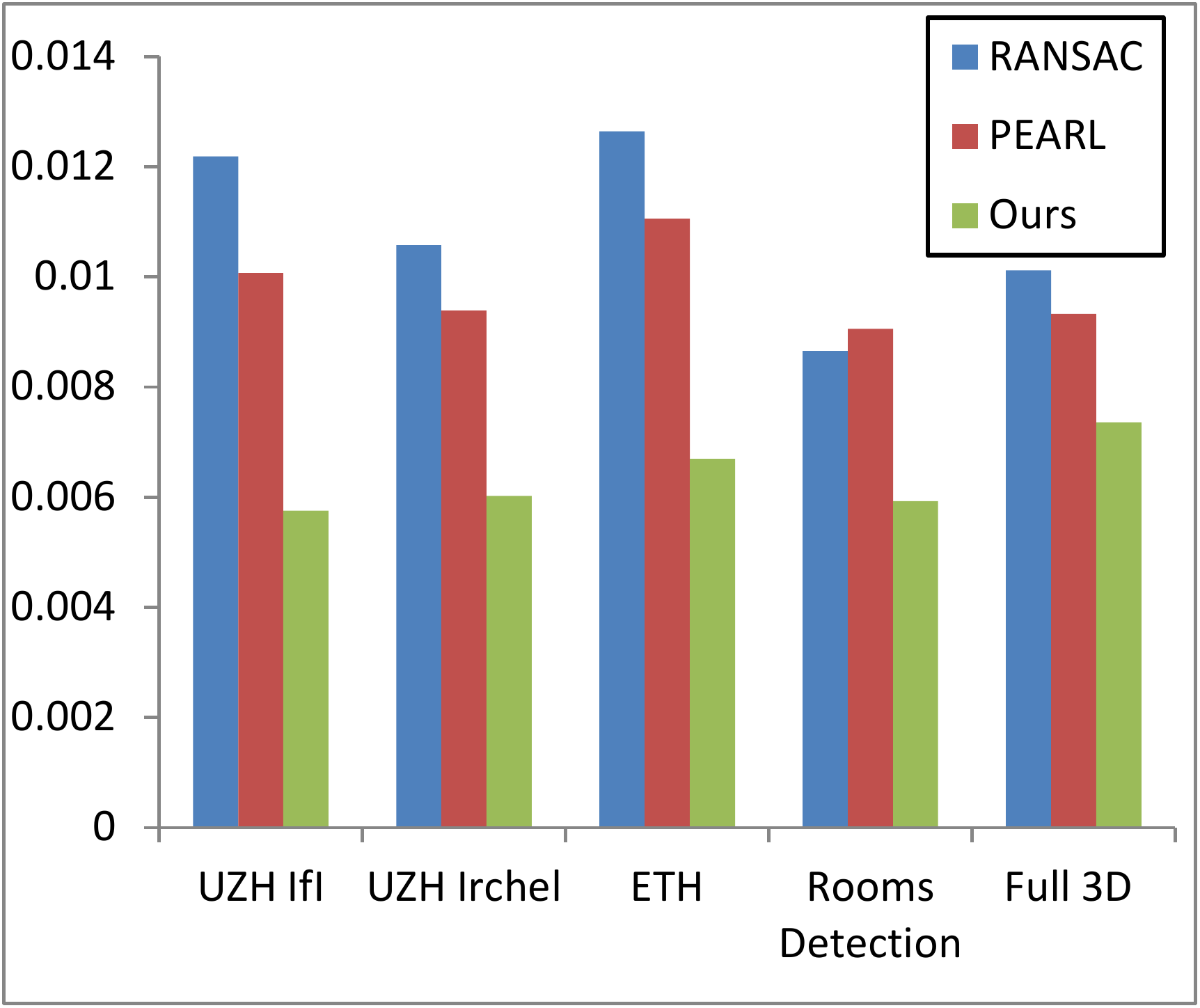}}
		\subfigure[$e_q$] {\includegraphics[width=1in]{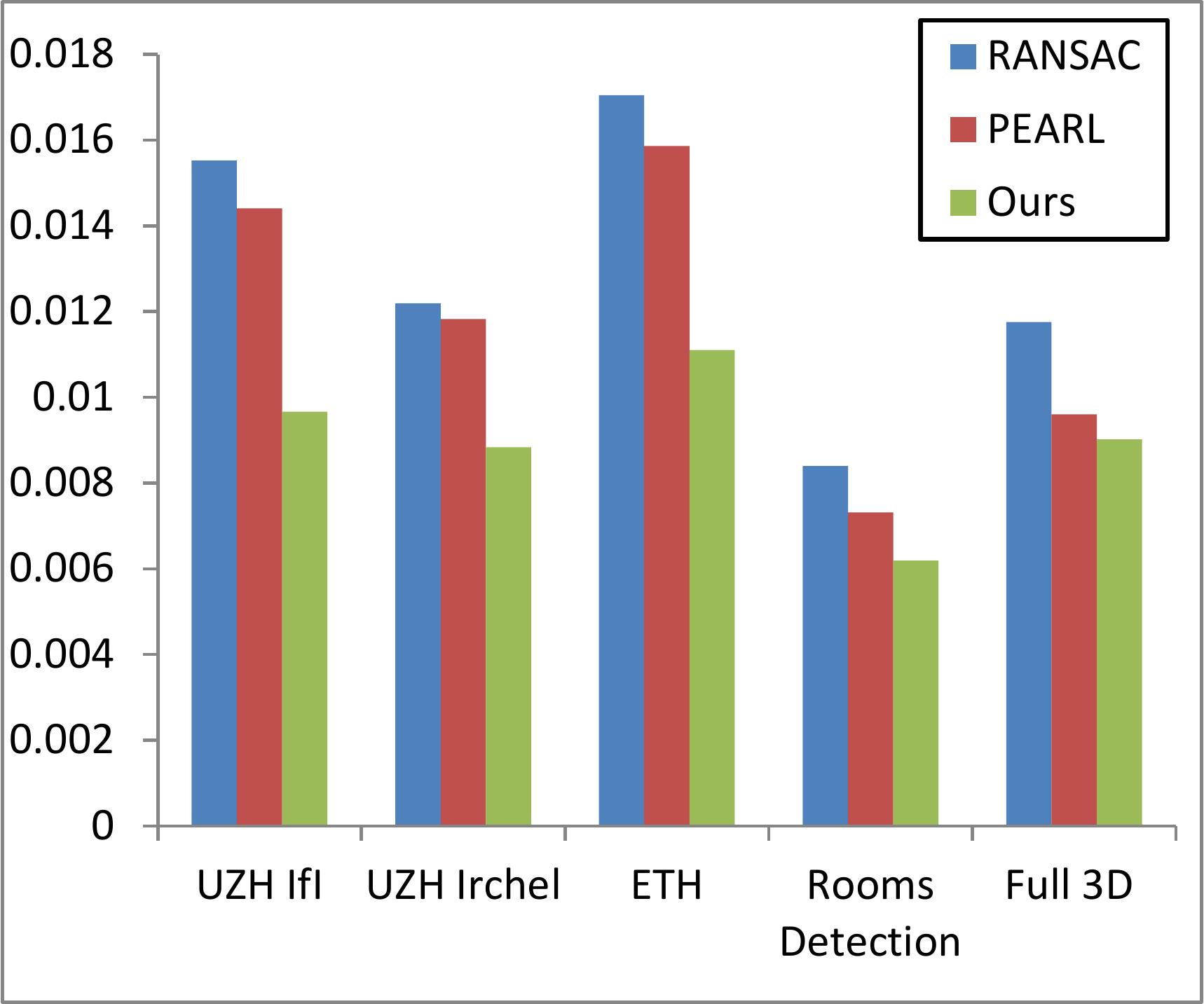}}
		\subfigure[Time (s)] {\includegraphics[width=1in]{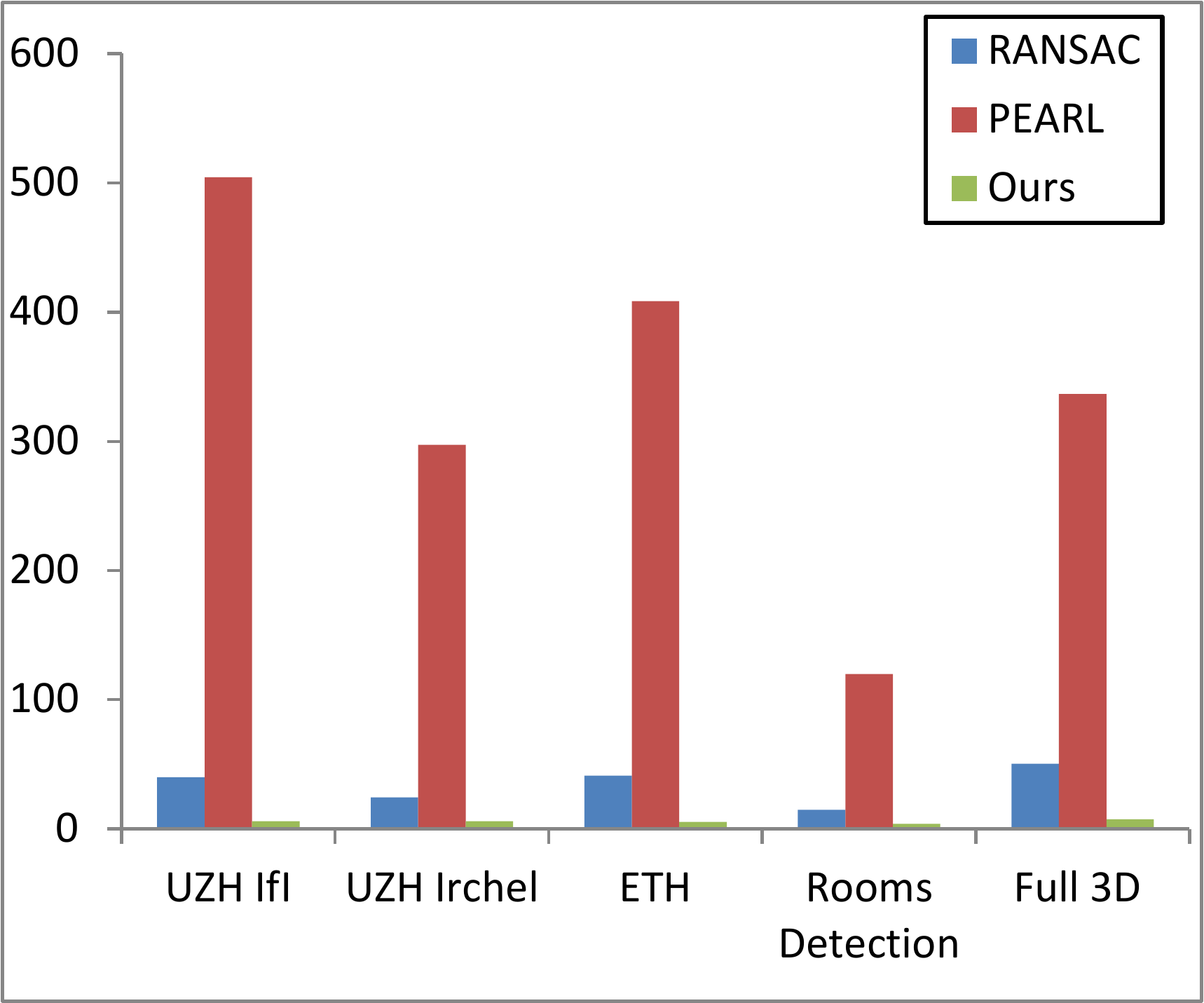}}
		\caption{Comparison results on indoor-point cloud data.}
		\label{fig:indoor_results}
	\end{figure}
	
\begin{figure}[htb]
		\centering
		\includegraphics[width=\linewidth, height=3.3in]{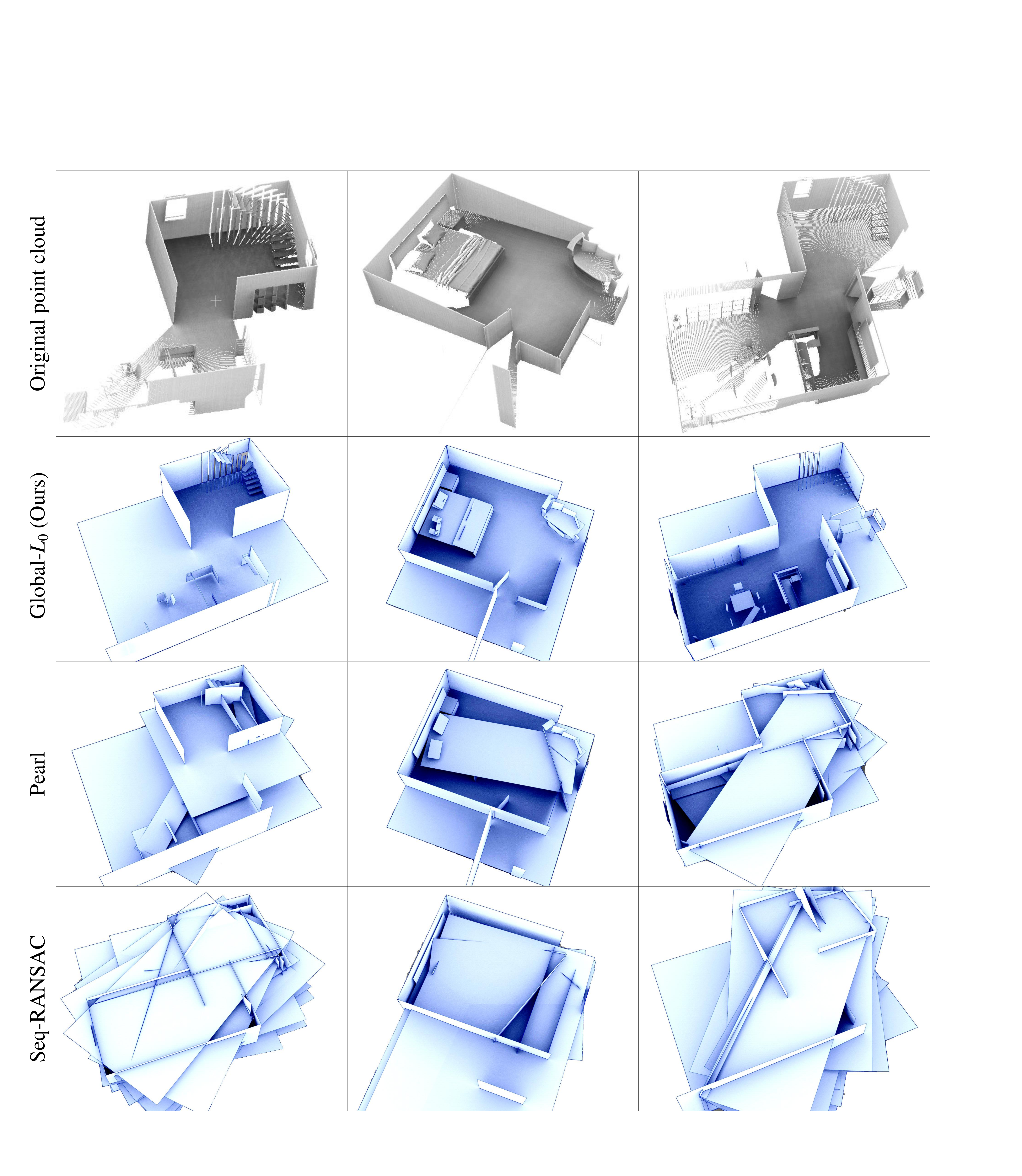}
		\caption{Plane reconstruction results of indoor point clouds.}
		\label{fig:indoor}
	\end{figure}

	The quantitative comparison results are shown in Fig.~\ref{fig:indoor_results} and some typical reconstruction results are shown in Fig.~\ref{fig:indoor}. Owing to the lack of ground truth, only $e_p$ and $e_q$ are evaluated here. Nevertheless, we can easily observe that our method is superior to the other two methods. PEARL has a high computational complexity and has difficulty maintaining small-scale planar structures. RANSAC-based methods tend to merge unconnected planes, making it difficult to maintain the edges of local structures. The proposed method not only maintains the local structure, it also guarantees the global constraints. Therefore, we achieve the best results.
	
	\section{Conclusion}
	
	In this paper, we presented a fast algorithm for RCPR problems. Our algorithm was based on a constraint model, requiring less prior knowledge compared to traditional RCPR algorithms. We showed that, by considering the constraints, plane reconstruction results could significantly improved, especially for data with high-level outliers and noise. Furthermore, our algorithm is efficient, between 3 and 1,000-times faster than the existing RCPR algorithms. 
	
	
	{\small
		\bibliographystyle{ieee}
		\bibliography{regular_plane}
	}
	
\end{document}